\documentclass[conference]{IEEEtran}
\usepackage[utf8]{inputenc}
\usepackage[hidelinks]{hyperref}
\usepackage{graphicx}
\usepackage{enumitem}
\usepackage{xcolor}
\usepackage{tikz}
\usetikzlibrary{patterns}
\usepackage{colortbl}
\usepackage{pgfplots}
\usepackage{pgfplotstable}
\pgfplotsset{compat=1.17}
\usepackage[
    giveninits=true,
    minnames=6,
    maxnames=6,
    isbn=false,
    url=false,
    doi=false
]{biblatex}
\title{Dynamic DNNs and Runtime Management for Efficient Inference on Mobile/Embedded Devices \\
{\footnotesize DATE PhD Forum 2024}\\[-1.0ex]
}
\author{\IEEEauthorblockN{Lei Xun}
\IEEEauthorblockA{\textit{School of Electronics \& Computer Science} \\
\textit{University of Southampton}, 
UK \\
l.xun@soton.ac.uk \\[-4.0ex]}
\and
\IEEEauthorblockN{\textbf{Advisor}: Jonathon Hare, Geoff V. Merrett}
\IEEEauthorblockA{\textit{School of Electronics \& Computer Science} \\
\textit{University of Southampton}, 
UK \\
\{jsh2, gvm\}@ecs.soton.ac.uk\\[-4.0ex]}}

\begin{document}

\maketitle

\section{Introduction}
Deep neural network (DNN) inference is increasingly being executed on mobile and embedded platforms \cite{cai2022enable, zhao2022survey} due to several key advantages in latency \cite{wu2019machine, yu2021automated}, privacy \cite{dai2019machine} and always-on availability \cite{wu2019machine, venieris2023nawq}. However, efficient DNN deployment on mobile and embedded platforms is challenging due to limited computing resources \cite{sze2017hardware, almeida2021smart}. Although many hardware accelerators \cite{han2016eie, chen2016eyeriss, safarpour2021high, safarpour2021low, moss2022ultra} and static model compression methods \cite{yang2018netadapt, he2018amc} were proposed by previous works, at system runtime, multiple applications are typically executed concurrently and compete for hardware resources. This raises two main challenges:

\begin{itemize}
\item \textbf{Runtime Hardware Availability:} Modern System-on-Chips (SoCs), comprising CPUs, GPUs, and NPUs, face challenges due to the varying availability of hardware resources at runtime. This fluctuation arises from different core combinations and changes in voltage and clock frequencies. While static model compression can initially optimize DNN models to fit on targeted hardware and meet performance goals, the issue is the unpredictable availability of these resources during runtime. Different and dynamic workloads on SoCs make it hard to consistently meet performance targets, as the hardware resources available to DNN models change and are unknown during the initial compression \cite{xun2020optimising, hoffmann2020embodied, bai2021batchquant, bai2022automated}.

\item \textbf{Runtime Application Variability:} A single DNN model, such as large language models (LLMs), can serve as the backbone for various applications like translation, text generation, and ChatBot, each requiring different performance trade-offs. For example, a ChatBot needs LLMs to have low latency for quick responses, whereas translation and text generation need LLMs to focus on accuracy. These performance targets can also change through user settings/preferences at runtime, posing a significant challenge in the design stage. The current solution of using multiple static models with different performance trade-offs is not feasible for mobile and embedded platforms due to limited memory resources. An ideal approach would be a single, adaptable DNN model that dynamically adjusts its performance trade-offs to meet the specific requirements of each application and user at runtime.
\end{itemize}

\section{Runtime System-level Performance Trade-off Management}

\begin{figure}[h]
\begin{center}
   \vspace{-6mm}
   \includegraphics[width=1.0\columnwidth]{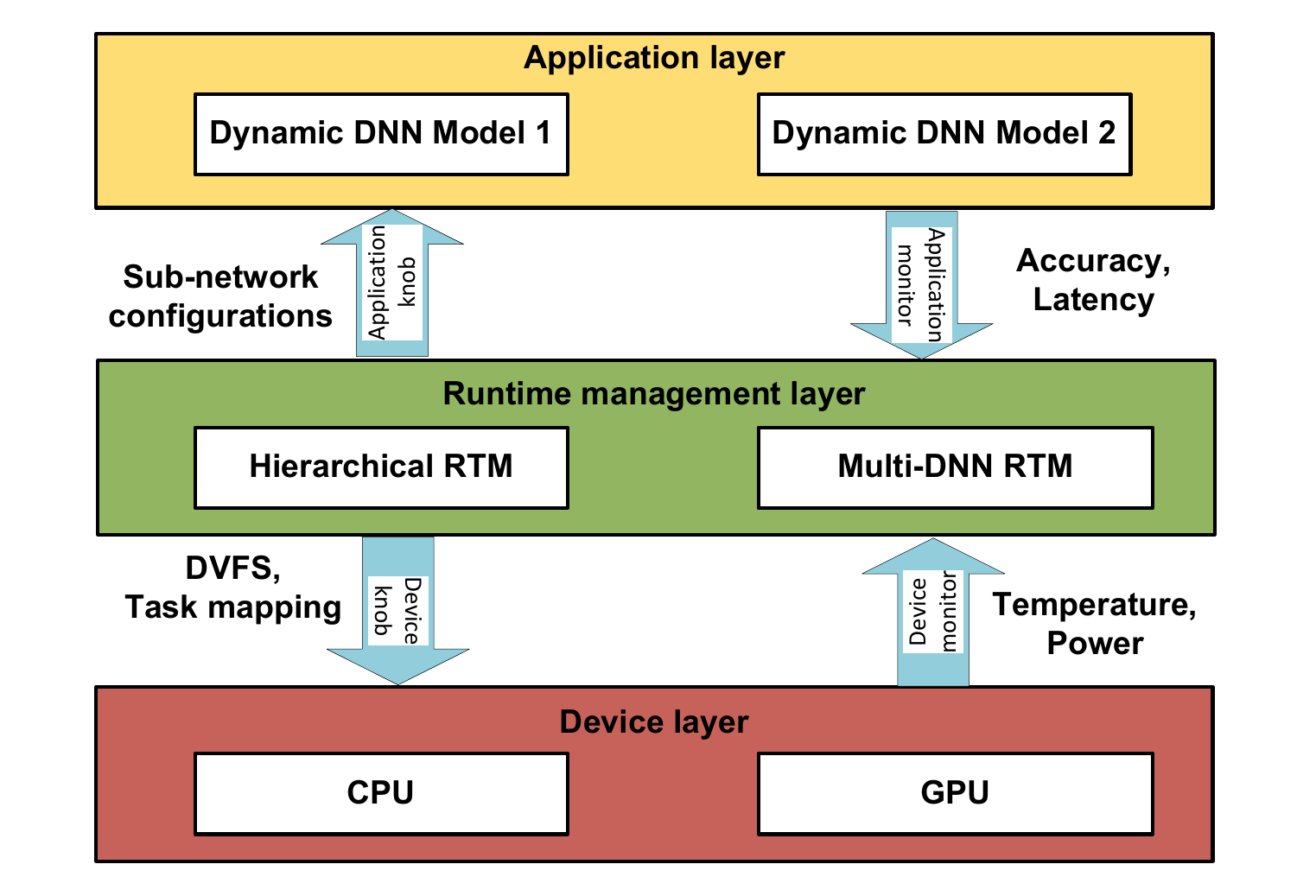}
    \vspace{-6mm}
   \caption{The high-level diagram of the proposed runtime system, which contains three abstract layers that are connected through knobs and monitors.}
   \vspace{-5mm}
\label{Fig: RTM}
\end{center}
\end{figure}

Previous works have addressed aforementioned challenges through dynamic neural networks that contain sub-networks with different performance trade-offs \cite{tann2016runtime, reform, yu2018slimmable, yu2019universally, yu2019autoslim, yang2021mutualnet} or runtime hardware resource management \cite{reddy2017inter, basireddy2019adamd, singh2017energy, singh2019collaborative, xu2021co}. 

We proposed a combined method \cite{xun2019incremental, xun2020optimising}, a system was developed for DNN performance trade-off management, combining the runtime trade-off opportunities in both algorithms and hardware (Fig \ref{Fig: RTM}). The runtime system contains three abstract layers which are connected through knobs and monitors. Application layers contain multiple concurrent dynamic neural networks. The device layer is mobile and embedded heterogeneous SoCs. The runtime management layer is the highest-level layer with central control algorithms for tuning both application knobs (i.e. dynamic DNN sub-networks), and device knobs (e.g. DVFS and task mapping) to meet dynamically changing application performance targets and hardware constraints which are both monitored in real-time. The key contribution of our works \cite{xun2019incremental, xun2020optimising, tinyml, xun2022dynamic, WNADA} is system-level performance trade-off management, the application layer and runtime management layer are co-designed based on the characterisation of the underlay hardware platforms to boost the space and granularity of performance trade-offs. This is different to the previous works which only focus on standalone system components while isolating others.

\section{Dynamic Super-networks}

\begin{figure}[h]
\vspace{-3mm}
\includegraphics[width=1\columnwidth]{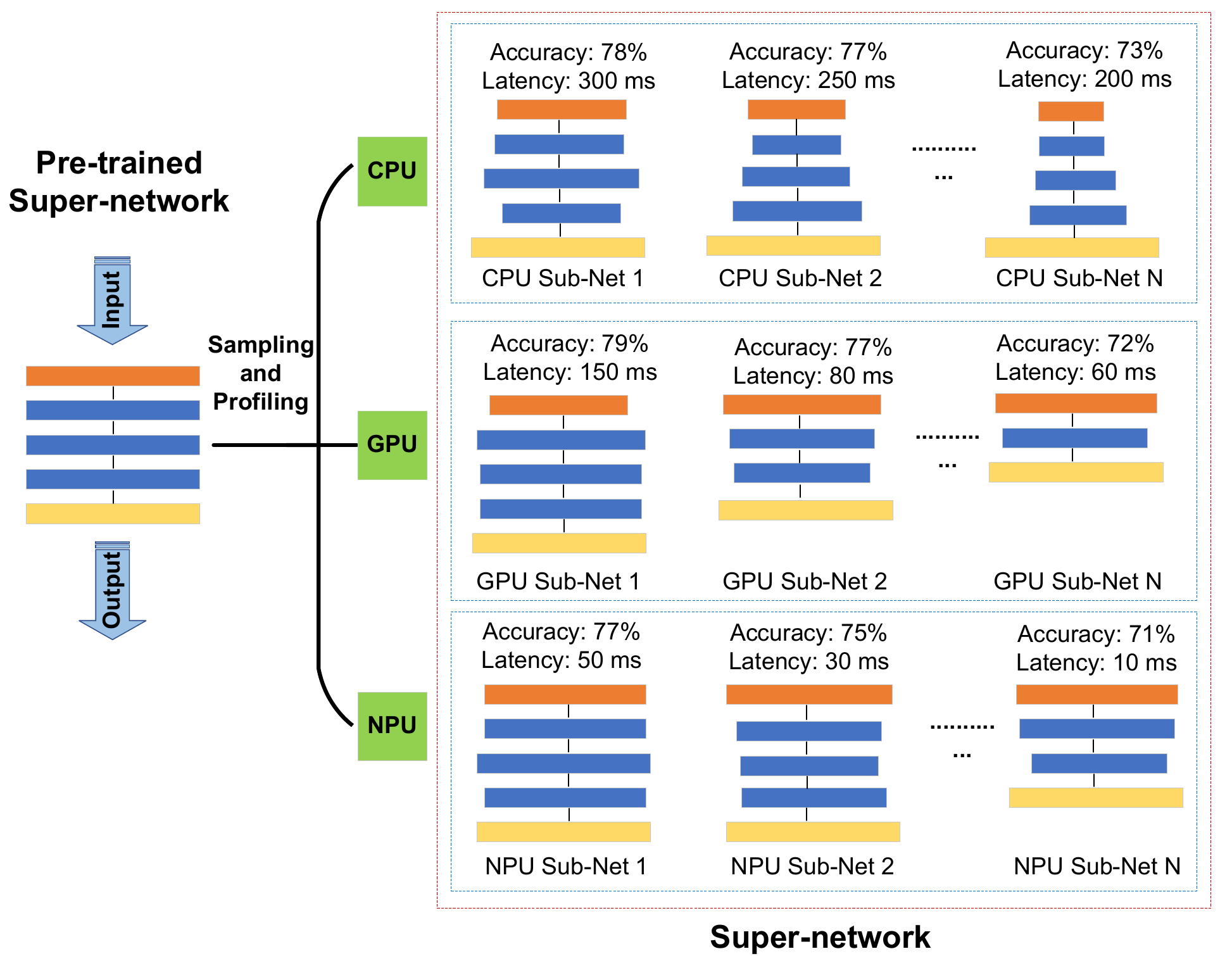}
\centering
\vspace{-3mm}
\caption{Dynamic Super-network. It samples and combines different efficient sub-network libraries from backbone super-networks for all heterogeneous cores, and to build dynamic neural networks without training.}
\vspace{-3mm}
\label{Fig: Dynamic Super-network}
\end{figure}

We co-designed novel dynamic neural networks to maximise system-level performance and energy efficiency. We identified three common issues with all previous dynamic neural networks: (1) significant training time cost, (2) conflict with the SOTA neural architecture search (NAS) pipeline, and (3) inference inefficiency on heterogeneous hardware. 

To address these problems, we proposed Dynamic Super-network \cite{dynamic-ofa, parry2021dynamic}, as shown in Fig \ref{Fig: Dynamic Super-network}, which directly samples efficient sub-networks on (near) the performance trade-off Pareto-front from the backbone super-networks to create a library and build dynamic neural networks directly without any training process (i.e. only sampling and performance profiling). The sampling and profiling process is repeated for different heterogeneous cores (e.g. CPU, GPU, NPU) on SoCs, this is because the most efficient DNN model architectures for different cores are different, and often contrary \cite{cai2018proxylessnas}. Super-networks have the flexibility to generate very different sub-network model architectures, whereas previous dynamic neural networks can scale either layer-wise \cite{teerapittayanon2016branchynet, wang2018skipnet} or channel-wise \cite{tann2016runtime, reform, yu2018slimmable, yu2019universally, yu2019autoslim, yang2021mutualnet}. In the end, we obtained different libraries of efficient sub-networks for all heterogeneous cores using only one set of model weights (i.e. the weights of the full super-network) which are stored using on-chip memory. Each library has multiple sub-networks with different performance trade-offs, which all share their weights with each other, and with sub-networks in other libraries, as well as the super-network. Each library is efficient on the corresponding cores, and when moving the DNN model between cores (e.g. from GPU to CPU), a different sub-network from the target hardware (i.e. CPU) model library is selected, and this is done by applying masks to the backbone super-network so it can be partially executed, this act as model/weights selection from the backbone model rather than full model/weights switching. 

\begin{figure}[t]
\begin{center}
   \vspace{-5mm}
   \includegraphics[width=1\columnwidth]{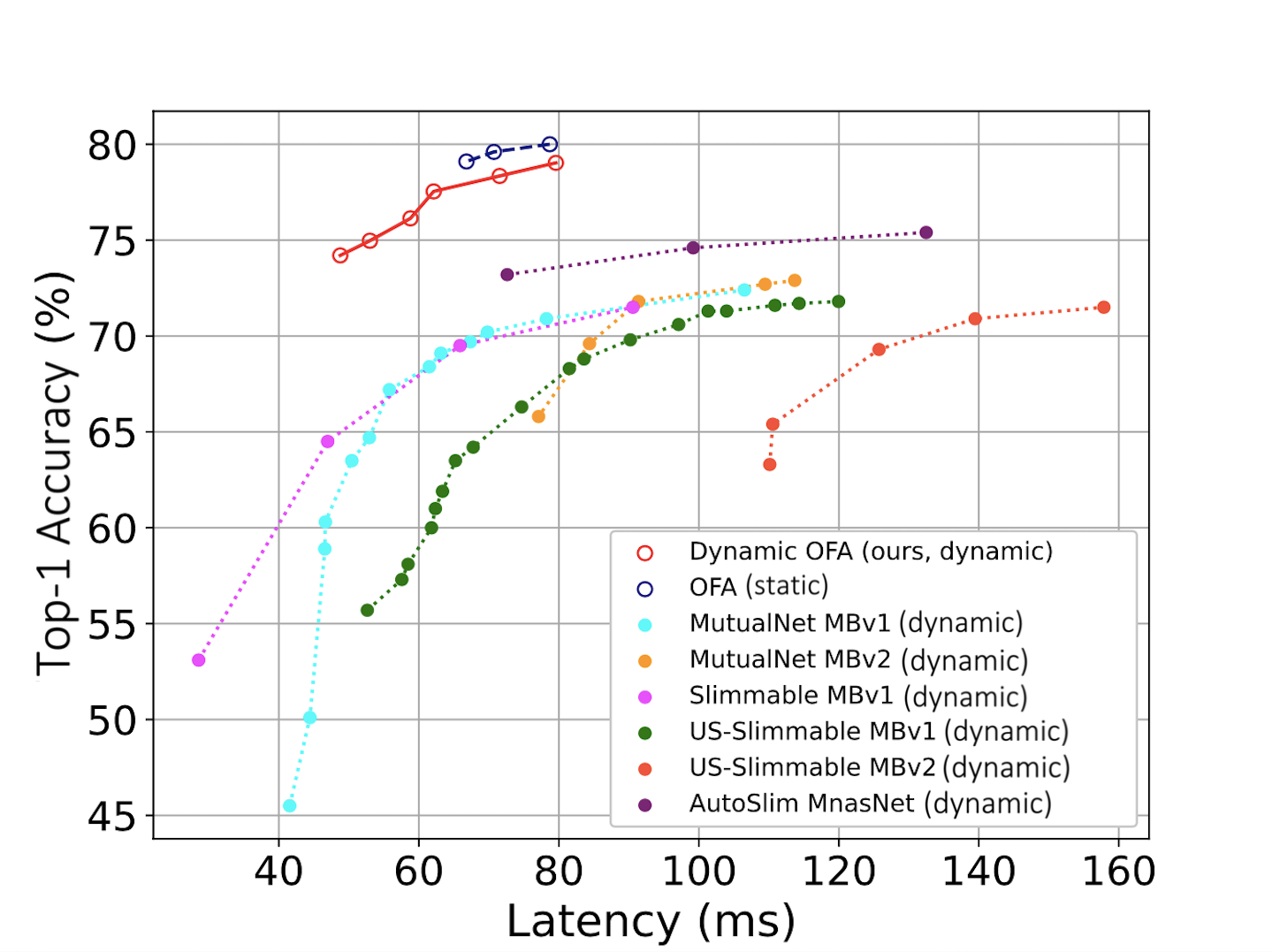}
   \vspace{-5mm}
   \caption{The performance trade-offs between ImageNet Top-1 accuracy and latency of our Dynamic OFA model \cite{dynamic-ofa} on the GPU of Jetson Xavier NX platform, comparing against SOTA static OFA backbone model \cite{cai2019once} and dynamic DNN models \cite{yu2018slimmable,yu2019universally, yu2019autoslim, yang2021mutualnet}. Dynamic OFA model is 2.4x faster (at similar accuracy) or has 5.1\% higher Top-1 ImageNet accuracy (at similar latency) than AutoSlim-MnasNet \cite{yu2019autoslim}.}
\label{Fig: Dynamic OFA Accuracy_vs_Latency_GPU}
\vspace{-5mm}
\end{center}
\end{figure}

We illustrated the Dynamic Super-network approach through a dynamic version of `once-for-all network \cite{cai2019once}' (namely Dynamic-OFA), which can scale the ConvNet architecture to fit heterogeneous computing resources efficiently \cite{dynamic-ofa} and has good generalisation for different model architectures such as Transformer \cite{parry2021dynamic}. As shown in Fig \ref{Fig: Dynamic OFA Accuracy_vs_Latency_GPU}, compared to the SOTA dynamic neural networks \cite{yu2018slimmable,yu2019universally, yu2019autoslim, yang2021mutualnet}, our experimental results using ImageNet on the GPU of Jetson Xavier NX show that the Dynamic-OFA is 2.4x faster for similar ImageNet Top-1 accuracy, or 5.1\% higher accuracy at similar latency. We also see a similar level of improvement on the CPU. Dynamic OFA has a slightly lower accuracy than its fine-tuned static versions (i.e. individual models with separate weights), but these models have significant memory overhead and runtime switching costs.

\section{Runtime Resource Management for Dynamic Super-networks}
We illustrated the runtime management approach through a hierarchical runtime resource manager that tunes both dynamic neural networks and DVFS at runtime to meet the hardware constraints (e.g. power consumption), and algorithm performance targets (e.g. accuracy, latency). Compared with the Linux DVFS governor schedutil, our runtime approach achieves up to a 19\% energy reduction and a 9\% latency reduction in single model deployment scenario, and an 89\% energy reduction and a 23\% latency reduction in a two concurrent model deployment scenario.

\section*{Acknowledgments}
These works were supported by the Engineering and Physical Sciences Research Council (EPSRC) under Grant EP/S030069/1. Experimental data can be found at: https://doi.org/10.5258/SOTON/D1804. Code is available open-source at https://github.com/UoS-EEC/DynamicOFA.

\cleardoublepage

\renewcommand*{\bibfont}{\footnotesize}
\printbibliography

@inproceedings{xun2019incremental,
  title={{Incremental Training and Group Convolution Pruning for Runtime DNN Performance Scaling on Heterogeneous Embedded Platforms}},
  author={Xun, Lei and Tran-Thanh, Long and Al-Hashimi, Bashir M and Merrett, Geoff V},
  booktitle={ACM/IEEE 1st Workshop on Machine Learning for CAD (MLCAD)},
  year={2019},
}

@inproceedings{parry2021dynamic,
  title={{Dynamic Transformer for Efficient Machine Translation on Embedded Devices}},
  author={Parry, Hishan and Xun, Lei and Sabet, Amin and Bi, Jia and Hare, Jonathon and Merrett, Geoff V},
  booktitle={ACM/IEEE 3rd Workshop on Machine Learning for CAD (MLCAD)},
  year={2021},
}

@inproceedings{xun2020optimising,
  title={{Optimising Resource Management for Embedded Machine Learning}},
  author={Xun, Lei and Tran-Thanh, Long and Al-Hashimi, Bashir M and Merrett, Geoff V},
  booktitle={Design, Automation and Test in Europe Conference (DATE)},
  year={2020},
}

@inproceedings{tinyml,
  title={{Runtime DNN Performance Scaling through Resource Management on Heterogeneous Embedded Platforms}},
  author={Xun, Lei and Al-Hashimi, Bashir M and Hare, Johnathan and Merrett, Geoff V},
  booktitle={tinyML EMEA Technical Forum},
  year={2021},
}

@inproceedings{dynamic-ofa,
  title={{Dynamic-OFA: Runtime DNN Architecture Switching for Performance Scaling on Heterogeneous Embedded Platforms}},
  author={Lou, Wei and Xun, Lei and Sabet, Amin and Bi, Jia and Hare, Jonathon and Merrett, Geoff V},
  booktitle={Conference on Computer Vision and Pattern Recognition (CVPR) Workshops},
  year={2021}
}

@article{safarpour2021high,
  title={{A High-Level Approach for Energy Efficiency Improvement of FPGAs by Voltage Trimming}},
  author={Safarpour, Mehdi and Xun, Lei and Merrett, Geoff V and Silv{\'e}n, Olli},
  journal={IEEE Transactions on Computer-Aided Design of Integrated Circuits and Systems (TCAD)},
  year={2021},
  publisher={IEEE}
}

@inproceedings{safarpour2021low,
  title={{Low-Voltage Energy Efficient Neural Inference by Leveraging Fault Detection Techniques}},
  author={Safarpour, Mehdi and Deng, Tommy Z and Massingham, John and Xun, Lei and Sabokrou, Mohammad and Silv{\'e}n, Olli},
  booktitle={Nordic Circuits and Systems Conference (NorCAS)},
  year={2021},
}

@inproceedings{xun2022dynamic,
  title={{Dynamic DNNs Meet Runtime Resource Management on Mobile and Embedded Platforms}},
  author={Xun, Lei and Al-Hashimi, Bashir M and Hare, Jonathon and Merrett, Geoff V},
  booktitle={UK Mobile, Wearable and Ubiquitous Systems Research Symposium (MobiUK)},
  year={2022}
}

@inproceedings{moss2022ultra,
  title={{Ultra-low Power DNN Accelerators for IoT: Resource Characterization of the MAX78000}},
  author={Moss, Arthur and Lee, Hyunjong and Xun, Lei and Min, Chulhong and Kawsar, Fahim and Montanari, Alessandro},
  booktitle={SenSys Conference 4th Workshop on AIChallengeIoT},
  year={2022}
}

@inproceedings{WNADA,
  title={{Dynamic DNNs Meet Runtime Resource Management for Efficient Heterogeneous Computing}},
  author={Xun, Lei and Hare, Johnathan and Merrett, Geoff V},
  booktitle={Workshop on Novel Architecture and Novel Design Automation (NANDA)},
  year={2023},
}

@inproceedings{he2018amc,
  title={{AMC: AutoML for Model Compression and Acceleration on Mobile Devices}},
  author={He, Yihui and Lin, Ji and Liu, Zhijian and Wang, Hanrui and Li, Li-Jia and Han, Song},
  booktitle={European Conference on Computer Vision (ECCV)},
  year={2018}
}

@inproceedings{yang2018netadapt,
  title={{NetAdapt: Platform-Aware Neural Network Adaptation for Mobile Applications}},
  author={Yang, Tien-Ju and Howard, Andrew and Chen, Bo and Zhang, Xiao and Go, Alec and Sandler, Mark and Sze, Vivienne and Adam, Hartwig},
  booktitle={European Conference on Computer Vision (ECCV)},
  year={2018}
}

@inproceedings{wang2018skipnet,
  title={{SkipNet: Learning Dynamic Routing in Convolutional Networks}},
  author={Wang, Xin and Yu, Fisher and Dou, Zi-Yi and Darrell, Trevor and Gonzalez, Joseph E},
  booktitle={European Conference on Computer Vision (ECCV)},
  year={2018}
}

@inproceedings{yu2019universally,
  title={{Universally Slimmable Networks and Improved Training Techniques}},
  author={Yu, Jiahui and Huang, Thomas S},
  booktitle={International Conference on Computer Vision (ICCV)},
  year={2019}
}

@inproceedings{cai2018proxylessnas,
  title={{ProxylessNAS: Direct Neural Architecture Search on Target Task and Hardware}},
  author={Cai, Han and Zhu, Ligeng and Han, Song},
  booktitle={International Conference on Learning Representations (ICLR)},
  year={2019}
}

@inproceedings{yu2018slimmable,
  title={{Slimmable Neural Networks}},
  author={Yu, Jiahui and Yang, Linjie and Xu, Ning and Yang, Jianchao and Huang, Thomas},
  booktitle={International Conference on Learning Representations (ICLR)},
  year={2019}
}

@inproceedings{cai2019once,
  title={{Once-for-All: Train One Network and Specialize it for Efficient Deployment}},
  author={Cai, Han and Gan, Chuang and Wang, Tianzhe and Zhang, Zhekai and Han, Song},
  booktitle={International Conference on Learning Representations (ICLR)},
  year={2020}
}

@inproceedings{bai2021batchquant,
  title={{BatchQuant: Quantized-for-all Architecture Search with Robust Quantizer}},
  author={Bai, Haoping and Cao, Meng and Huang, Ping and Shan, Jiulong},
  booktitle={Advances in Neural Information Processing Systems (NeurIPS)},
  year={2021}
}

@inproceedings{reform,
  title={{ReForm: Static and Dynamic Resource-Aware DNN Reconfiguration Framework for Mobile Device}},
  author={Xu, Zirui and Yu, Fuxun and Liu, Chenchen and Chen, Xiang},
  booktitle={Design Automation Conference (DAC)},
  year={2019},
}

@inproceedings{yu2021automated,
  title={{Automated Runtime-Aware Scheduling for Multi-Tenant DNN Inference on GPU}},
  author={Yu, Fuxun and Bray, Shawn and Wang, Di and Shangguan, Longfei and Tang, Xulong and Liu, Chenchen and Chen, Xiang},
  booktitle={International Conference On Computer-Aided Design (ICCAD)},
  year={2021},
}

@inproceedings{tann2016runtime,
  title={{Runtime Configurable Deep Neural Networks for Energy-Accuracy Trade-off}},
  author={Tann, Hokchhay and Hashemi, Soheil and Bahar, R and Reda, Sherief},
  booktitle={International Conference on Hardware/Software Codesign and System Synthesis (CODES+ISSS)},
  year={2016},
}

@inproceedings{han2016eie,
  title={{EIE: Efficient Inference Engine on Compressed Deep Neural Network}},
  author={Han, Song and Liu, Xingyu and Mao, Huizi and Pu, Jing and Pedram, Ardavan and Horowitz, Mark A and Dally, William J},
  booktitle={International Symposium on Computer Architecture (ISCA)},
  year={2016},
}

@inproceedings{wu2019machine,
  title={{Machine Learning at Facebook: Understanding Inference at the Edge}},
  author={Wu, Carole-Jean and Brooks, David and Chen, Kevin and Chen, Douglas and Choudhury, Sy and Dukhan, Marat and Hazelwood, Kim and Isaac, Eldad and Jia, Yangqing and Jia, Bill and others},
  booktitle={International Symposium on High Performance Computer Architecture (HPCA)},
  year={2019},
}

@article{hoffmann2020embodied,
  title={{Embodied Self-Aware Computing Systems}},
  author={Hoffmann, Henry and Jantsch, Axel and Dutt, Nikil D},
  journal={Proceedings of the IEEE},
  volume={108},
  number={7},
  pages={1027--1046},
  year={2020},
  publisher={IEEE}
}

@article{zhao2022survey,
  title={{A Survey of Deep Learning on Mobile Devices: Applications, Optimizations, Challenges, and Research Opportunities}},
  author={Zhao, Tianming and Xie, Yucheng and Wang, Yan and Cheng, Jerry and Guo, Xiaonan and Hu, Bin and Chen, Yingying},
  journal={Proceedings of the IEEE},
  volume={110},
  number={3},
  pages={334--354},
  year={2022},
  publisher={IEEE}
}

@article{basireddy2019adamd,
  title={{AdaMD: Adaptive Mapping and DVFS for Energy-Efficient Heterogeneous Multicores}},
  author={Basireddy, Karunakar R and Singh, Amit Kumar and Al-Hashimi, Bashir M and Merrett, Geoff V},
  journal={IEEE Transactions on Computer-Aided Design of Integrated Circuits and Systems (TCAD)},
  year={2019},
  publisher={IEEE}
}

@article{singh2017energy,
  title={{Energy-Efficient Run-Time Mapping and Thread Partitioning of Concurrent OpenCL Applications on CPU-GPU MPSoCs}},
  author={Singh, Amit Kumar and Prakash, Alok and Basireddy, Karunakar Reddy and Merrett, Geoff V and Al-Hashimi, Bashir M},
  journal={ACM Transactions on Embedded Computing Systems (TECS)},
  volume={16},
  number={5s},
  pages={1--22},
  year={2017},
  publisher={ACM New York, NY, USA}
}

@article{singh2019collaborative,
  title={{Collaborative Adaptation for Energy-Efficient Heterogeneous Mobile SoCs}},
  author={Singh, Amit Kumar and Basireddy, Karunakar Reddy and Prakash, Alok and Merrett, Geoff V and Al-Hashimi, Bashir M},
  journal={IEEE Transactions on Computers (TC)},
  volume={69},
  number={2},
  pages={185--197},
  year={2019},
  publisher={IEEE}
}

@article{bai2022automated,
  title={{Automated Customization of On-Device Inference for Quality-of-Experience Enhancement}},
  author={Bai, Yang and Chen, Lixing and Ren, Shaolei and Xu, Jie},
  journal={IEEE Transactions on Computers (TC)},
  year={2022},
  publisher={IEEE}
}

@article{chen2016eyeriss,
  title={{Eyeriss: An Energy-Efficient Reconfigurable Accelerator for Deep Convolutional Neural Networks}},
  author={Chen, Yu-Hsin and Krishna, Tushar and Emer, Joel S and Sze, Vivienne},
  journal={IEEE Journal of Solid-State Circuits},
  volume={52},
  number={1},
  pages={127--138},
  year={2016},
  publisher={IEEE}
}

@article{yang2021mutualnet,
  title={{MutualNet: Adaptive ConvNet via Mutual Learning from Different Model Configurations}},
  author={Yang, Taojiannan and Zhu, Sijie and Mendieta, Matias and Wang, Pu and Balakrishnan, Ravikumar and Lee, Minwoo and Han, Tao and Shah, Mubarak and Chen, Chen},
  journal={IEEE Transactions on Pattern Analysis and Machine Intelligence (TPAMI)},
  year={2021},
  publisher={IEEE}
}

@article{yu2019autoslim,
  title={{AutoSlim: Towards One-Shot Architecture Search for Channel Numbers}},
  author={Yu, Jiahui and Huang, Thomas},
  journal={arXiv preprint arXiv:1903.11728},
  year={2019}
}

@article{venieris2023nawq,
  title={{NAWQ-SR: A Hybrid-Precision NPU Engine for Efficient On-Device Super-Resolution}},
  author={Venieris, Stylianos I and Almeida, Mario and Lee, Royson and Lane, Nicholas D},
  journal={IEEE Transactions on Mobile Computing},
  year={2023},
  publisher={IEEE}
}

@inproceedings{teerapittayanon2016branchynet,
  title={{BranchyNet: Fast Inference via Early Exiting from Deep Neural Networks}},
  author={Teerapittayanon, Surat and McDanel, Bradley and Kung, Hsiang-Tsung},
  booktitle={International Conference on Pattern Recognition (ICPR)},
  year={2016},
}

\end{document}